\documentclass[runningheads]{llncs}
\usepackage{graphicx}
\usepackage{times}  
\usepackage{helvet}  
\usepackage{courier}  
\usepackage[hyphens]{url}  
\usepackage{graphicx} 
\usepackage{caption} 
\usepackage{url}
\usepackage[table]{xcolor} 
\usepackage{algorithm}
\usepackage{tabularx} 

\usepackage{amsmath}
\usepackage{graphicx}
\usepackage{multirow} 
\usepackage{xcolor}
\usepackage{amsmath}
\usepackage{algorithm}   
\usepackage{algpseudocode}
\usepackage{algorithm}
\usepackage{amssymb}
\usepackage{amsmath}  
\usepackage{graphicx}
\usepackage{booktabs, tabularx}
\newcolumntype{C}{>{\centering\arraybackslash}X}
\usepackage{multirow}
\usepackage{amsmath,amssymb}
\usepackage{times}
\usepackage{helvet}
\usepackage{courier}
\usepackage{xcolor}

\begin{document}
\title{DPAMixerSR: An Efficient Degradation-Pattern-Aware Model for Image Super-Resolution}
%
%
\author{Song-Li Wu\inst{1} \and
Haonan Jiang\inst{1} \and
Jixuan Fan\inst{1} \and
Yufei Huo\inst{1} \and \\
Chubin Zhang\inst{1} \and
Yansong Tang\thanks{Corresponding Author}\inst{1}
}

\institute{$^1$ Shenzhen International Graduate School, Tsinghua University  \\
\email{\{wsl24,jiang-hn24,fjx23,huoyf24,zcb24\}@mails.tsinghua.edu.cn, \\tang.yansong@sz.tsinghua.edu.cn} }
%
\maketitle              
\begin{abstract}
While content-adaptive schemes have delivered notable advances in image super-resolution (SR), existing approaches typically focus on texture complexity and ignore intrinsic degradation factors (e.g., blur kernels or noise patterns), leading to suboptimal computation allocation and reconstruction performance. To remedy this, we propose DPAMixerSR, a degradation-pattern-aware framework that enables efficient SR through adaptive sparse computation. We design a lightweight Perceptual Degradation Ranking (PDR) module partitions the image into severely and mildly degraded patches, which are routed to the Adaptive Sparse Processing (ASP) and a lightweight convolutional branch, respectively.
ASP performs structure-aligned, multi-scale sparse propagation and bidirectional refinement, while the convolutional branch enhances efficiency in mildly degraded regions. By coupling degradation-driven routing with structure-aligned sparse processing, DPAMixerSR establishes a self-regulating framework that dynamically balances computational efficiency and reconstruction fidelity. Extensive experiments on various SR tasks demonstrate that our DPAMixerSR achieves superior structural restoration and perceptual fidelity with markedly reduced computational overhead, providing a novel and scalable framework for degradation-aware, resource-efficient SR.

\end{abstract}

\begin{figure}[t]
  \centering
  \includegraphics[width=1\linewidth]{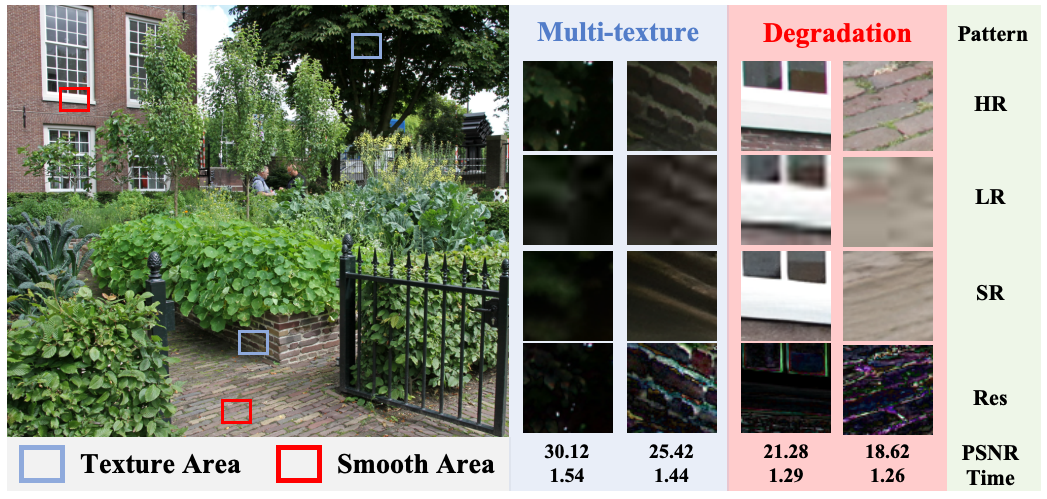}
  \setlength{\belowcaptionskip}{-6mm}
  \caption{Visual and quantitative comparison across regions. Texture-rich areas (\textcolor{blue}{blue box}) with mild degradation can be well restored by lightweight processing, whereas smooth areas (\textcolor{red}{red box}) with severe structural degradation require heavier computation. Conventional texture-based allocation overlooks degradation semantics, while our degradation-aware mechanism adaptively aligns computation with restoration fidelity.} 
  \label{motivation}
\end{figure}

\section{Introduction}
Aiming to reconstruct a high-resolution (HR) image from a low-resolution (LR) input, image super-resolution (SR) has been a long-standing yet inherently ill-posed problem in low-level vision~\cite{chen2022real}. Despite continuous progress in network architectures~\cite{liang2021swinir} and optimization objectives~\cite{wang2018esrgan,blau2018perception}, real-world SR remains challenging due to the complex, entangled degradations introduced by sensor noise, motion blur, and compression artifacts. These degradations are often unknown and spatially variant, making precise restoration highly dependent on accurate degradation modeling and structure-aware reconstruction.

Recent studies have embraced content-adaptive computation, where models dynamically allocate computational resources according to local texture complexity~\cite{ilhan2024resource}. This strategy assumes that regions with rich high-frequency details demand more intensive processing. While this is often effective, it overlooks another crucial factor: the intrinsic degradation characteristics of an image—such as blur kernels, noise intensity, or compression levels—which can also strongly influence restoration difficulty. As illustrated in Figure~\ref{motivation}, texture-dense regions (e.g., foliage, blue) with mild degradations can be effectively reconstructed even under lightweight processing, whereas visually smooth areas (e.g., pavement, bricks, red) suffering from severe corruption demand heavier computation yet often yield lower PSNR/SSIM. This observation highlights an important limitation: pixel-wise metrics and texture-driven routing can not fully capture true restoration difficulty, which results in suboptimal resource allocation—where computational effort is disproportionately spent on simple textures while regions with severe structural degradation are under-processed. Existing frameworks~\cite{zhou2025tsp,park2024graph} further amplify this limitation. Their rigid sparsity patterns often disrupt structural continuity by propagating features across semantically unrelated regions, resulting in artificial edges and spatial inconsistency. Moreover, unstable sparsity masks cause topological discontinuities during reconstruction, undermining coherence and perceptual quality.

To address these inefficiencies, we propose Degradation-Pattern-Aware Mixer Super-Resolution (DPAMixerSR), a novel framework that implements degradation-pattern-aware routing to guide adaptive computation. Rather than depending solely on local content complexity, DPAMixerSR dynamically routes computational effort according to localized degradation severity. At its core lies the Perceptual Degradation Ranking (PDR), which integrates contrastive representation learning with distilled knowledge from a high-fidelity quality assessor to hierarchically rank degradation intensity across spatial patches. This perceptual ranking establishes a principled basis for fine-grained and semantically consistent resource allocation. Building upon the PDR hierarchy, the Adaptive Sparse Processing (ASP) constructs semantically coherent propagation paths through sparsely activated and locally adaptive hierarchies. ASP routes heavily degraded regions into specialized reconstruction streams, while directing mildly degraded ones through lightweight convolutional pathways—achieving near-linear computational complexity without compromising structural fidelity. To stabilize multi-level feature propagation under diverse degradation conditions, we further introduce a Spatial Structure Adjustment (SSA) mechanism that adaptively calibrates hierarchical features along degradation gradients. The synergy of ASP and SSA forms a degradation-aware computational topology that harmonizes efficiency and fidelity, enabling DPAMixerSR to deliver superior perceptual quality with balanced computational cost.

In summary, our main contributions are threefold: 
\begin{itemize}  
\item We propose DPAMixerSR, an efficient and scalable SR framework that employs degradation-pattern-aware routing to adaptively address intrinsic degradation severity, an often-overlooked factor in SR.
\item We propose a Perceptual Degradation Ranking (PDR) for hierarchical evaluation of degradation severity, alongside an Adaptive Sparse Processing (ASP) module to enable structure-aligned feature propagation with dynamically adjustable sparsity.
\item Extensive experiments demonstrate DPAMixerSR achieves superior structural fidelity and perceptual realism under reduced computational budgets, establishing leading quality–efficiency trade-offs and scalability on lightweight and real-world SR tasks.
\end{itemize}


\section{Related work}

\noindent\textbf{Image Super-Resolution.}
While CNN-~\cite{wang2018esrgan} and Transformer-based~\cite{liang2021swinir} models have significantly advanced SR via hierarchical learning and long-range dependency modeling, their reliance on synthetic training data (e.g., bicubic downsampling) severely limits their robustness against complex, real-world degradations.

\noindent\textbf{Efficient Super-Resolution.}
To balance computational cost and quality, efficient SR employs content-aware allocation~\cite{wang2024camixersr} or lightweight architectures~\cite{hui2019lightweight}. Recently, many Mamba-based models~\cite{gu2024mamba,lei2024dvmsr} have emerged as linear-complexity alternatives to Transformers. However, their inherently 1D sequential nature restricts their adaptability to 2D local spatial structures.

\noindent\textbf{Real-World Image Super-Resolution.}
Conventional models~\cite{wang2018esrgan,blau2018perception} generalize poorly to real-world degradations. To address this, Real-ISR leverages diffusion generative priors~\cite{podell2023sdxl} (e.g., StableSR~\cite{wang2024exploiting}, DiffBIR~\cite{lin2024diffbir}, SeeSR~\cite{wu2024seesr}, and SUPIR~\cite{yu2024scaling}) to recover natural textures. Despite impressive generative capacity, these methods struggle to reconstruct fine structures due to the aggressive downsampling of the SD VAE~\cite{yi2025fine}. Furthermore, attempts to resolve this via larger backbones or multi-stage processes heavily inflate computational costs, limiting practical deployment.

\begin{figure*}[t]
  \centering
  \includegraphics[width=0.85\linewidth]{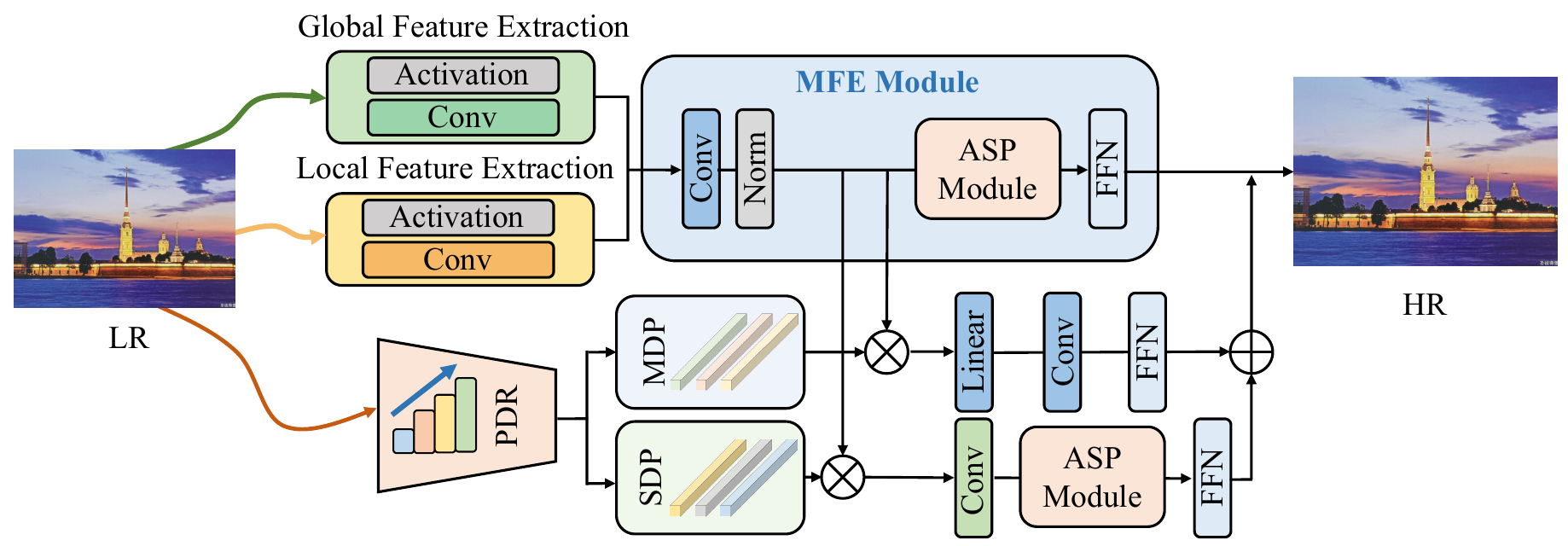}
  \caption{ DPAMixerSR comprises four key components: (1) Multi-scale Feature Extraction (MFE), which integrates hierarchical information for robust feature representation; (2) Perceptual Degradation Ranking (PDR), which dynamically assesses degradation severity to guide reconstruction; (3) Adaptive Sparse Processing (ASP) module, specialized for handling severely degraded patches (SDP); and (4) Convolution Branch, optimized for efficiently restoring mildly degraded patches (MDP).}
  \label{fig:overview}
\end{figure*}

\section{Methods}
\subsection{Overview}
To tackle the challenge of computational efficiency in real-world SR while preserving perceptual fidelity, we propose an efficient and scalable framework that dynamically allocates computation based on patch-level degradation severity. As shown in Figure~\ref{fig:overview}, our architecture consists of four key components: Multi-scale Feature Extraction (MFE),  Perceptual Degradation Ranking (PDR), Adaptive Sparse Processing (ASP) module and Convolution Branch. Given input \(I_{\mathrm{LR}}\), we extract deep features $\mathbf{FM}$ using a ConvNeXt backbone. A lightweight Perceptual Degradation Ranking (PDR) ranks each \(M\times M\) patch, dividing them into severe and mild groups. Severely degraded patches are processed by an Adaptive Sparse Processing (ASP) module with Spatial Structure Adjustment (SSA), while the others are handled by a lightweight convolutional branch. The outputs are then fused to produce \(I_{\mathrm{SR}}\). This adaptive routing mechanism ensures computational efficiency by focusing intensive processing only on the severely degraded regions.

\subsection{Perceptual Degradation Assessment Pipeline}
\label{subsec:pdap}
\begin{figure*}[t]
  \centering
  \includegraphics[width=0.85\linewidth]{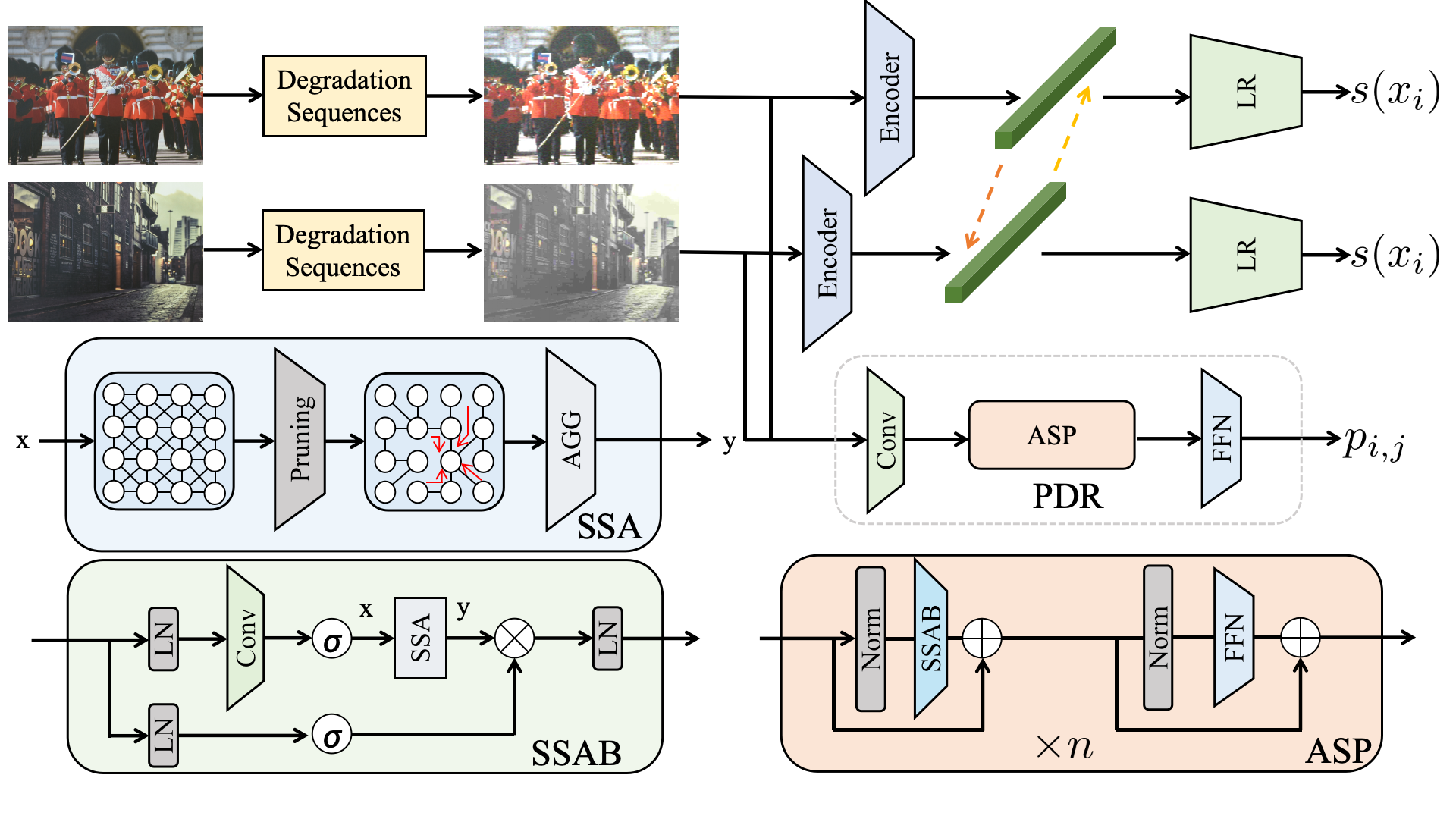}
  \caption{Our PDR model adopts a two-stage training scheme: a high-capacity PDS model first estimates degradation severity, and its outputs supervise the learning of a lightweight PDR model through ranking-aware objectives. To enhance degradation representation, the Spatial Structure Adjustment (SSA) module removes redundant features via SSAB blocks. In the ASP branch, SSAB blocks are combined with feed-forward network (FFN) layers and normalization (Norm) to enable adaptive restoration of complex regions. The aggregation (AGG) module performs bidirectional feature aggregation over structure-guided graphs, integrating local and contextual cues to improve reconstruction quality.} 
  \label{model9}
\end{figure*}

Accurate yet lightweight estimation of local degradation is essential for efficient routing, but compact models alone often struggle with complex degradations.
 To resolve this, we propose a two-stage score-to-rank scheme: a high-capacity Perceptual Degradation Score (PDS) model first predicts patch-wise degradation scores, which are then used to distill a lightweight Perceptual Degradation Ranking (PDR) model. This design ensures perceptual reliability and inference efficiency, enabling fast patch prioritization. The training pipeline consists of three modules: degradation sequence generation, multi-objective representation learning, and distillation-based ranking, as illustrated in Figure~\ref{model9}.

\subsubsection{Degradation Sequence Generation}
Real-world degradation modeling requires diverse training samples that mirror complex real-world degradations. We generate synthetic degradation sequences through stratified operator sampling, as shown in Algorithm~\ref{alg:degrade},
where $M_{\max}$ denotes the max length of degradation sequences and atomic operator \(D_{ij}\) denotes degradation operations, such as blurring and JPEG compression. This stochastic degradation strategy systematically samples diverse operator combinations and intensity levels, enabling broad coverage of degradation patterns while ensuring that the resulting distortions remain physically plausible and representative of real-world artifacts.
\begin{algorithm}[H]
  \caption{Degradation Sequence Generation}
  \label{alg:degrade}
  \begin{algorithmic}[1]
    \State Sample sequence length \(m \sim \mathcal{U}(1, M_{\max})\)
    \State Stratified sampling: select \(m\) distinct degradation types
    \State For each type, choose atomic operator \(D_{ij}\)
    \State Sample intensity levels \(l_{ij} \sim \mathcal{N}(0, \sigma^2)\)
    \State Apply operators in random order: \(I' = \mathcal{D}_{m} \circ \cdots \circ \mathcal{D}_1(I)\)
    \State \Return degradation sequence \(C_k\), degraded image \(I'\)
  \end{algorithmic}
\end{algorithm}
\subsubsection{Multi-Objective Representation Learning}
To capture degradation-aware and spatially sensitive features, we train a ConvNeXt encoder $f(\cdot)$ with three complementary objectives: a cross-image similarity loss $\mathcal{L}_1$ that aligns embeddings of images with identical degradation sequences, an intra-image diversity loss $\mathcal{L}_2$ that encourages variation across crops from the same degraded image, and a degradation discrimination loss $\mathcal{L}_3$ that separates features from different degradation sequences. Detailed formulations of $\mathcal{L}_1$, $\mathcal{L}_2$, and $\mathcal{L}_3$ are presented in the Appendix. When ground-truth scores from the PDS model are available, we incorporate a supervised regression loss $\mathcal{L}_{\mathrm{sup}} = \|r(f(I')) - y\|^2$. The overall training objective is:
\begin{equation}
\mathcal{L}_{\mathrm{total}} = \sum_{i=1}^3 \lambda_i \mathcal{L}_i + \mathbb{1}_{\mathrm{sup}}\lambda_4\mathcal{L}_{\mathrm{sup}},
\end{equation}
with ($\mathbb{1}_{\mathrm{sup}} = 1$) when supervised scores are available, and 0 otherwise. This multi-objective scheme shapes a degradation-sensitive feature space where both inter-patch discriminability and intra-image consistency are preserved, facilitating reliable relative ranking of degradation severity.

\subsubsection{Distillation-Based Ranking}

Directly deploying the high-capacity PDS model for patch-wise degradation estimation is impractical for real-time inference: absolute scores are noisy, scale-sensitive, and unreliable for routing decisions. 
Motivated by the relational nature of perceptual judgments~\cite{parducci2024perceptual}, we distill the PDS into a lightweight Perceptual Degradation Ranking (PDR) model, providing a robust and compact supervisory signal for patch prioritization. Formally, for any patch pair $(x_i, x_j)$, we define the corresponding ground-truth label as:

\begin{equation}
y_{ij} = \mathbb{I}\bigl(\mathrm{PDS}(x_i) > \mathrm{PDS}(x_j)\bigr), 
\quad
p_{ij} = \sigma\bigl(z_i^\top w - z_j^\top w\bigr),
\end{equation}
where $z_i$ and $z_j$ are the PDR embeddings and $\sigma(\cdot)$ denotes the sigmoid function. Enhanced with hard-negative mining to focus learning on ambiguous cases, we optimize the ranking model via cross-entropy loss:
\begin{equation}
\mathcal{L}_{\mathrm{rank}} = -\sum y_{ij} \log p_{ij} + (1 - y_{ij}) \log (1 - p_{ij}).
\end{equation}
The score-to-rank distillation scheme converts high-capacity perceptual knowledge into a compact, robust ranking signal, enabling PDR to not only prioritize patches with fine-grained degradation awareness but also dynamically allocate computation across the image—capturing intrinsic degradation hierarchies for efficient and scalable SR.

\subsection{Degradation-Pattern-Aware Mixing}
\label{subsec:dpamixer}

DPAMixer dynamically routes image regions to specialized branches based on estimated restoration difficulty, achieving a balance between efficiency and structural fidelity. Given input $X \in \mathbb{R}^{C \times H \times W}$, we extract texture features via pointwise convolution and geometric features via a $3 \times 3$ convolution with GELU activation. These are fused with degradation-aware context $C_w$ through an ASP module to form feature map $\mathrm{FM}$, from which we compute a deformation field $\Delta p = \tanh(f_{\mathrm{offsets}}(\mathrm{FM}))$ and a score map $s = \sigma(f_{\mathrm{score}}(\mathrm{FM}))$. $\Delta p$ guides content-aware warping of $X$ via bilinear sampling to obtain $\widetilde{X}$, while $s$ quantifies local degradation severity.
The warped features and scores are partitioned into non-overlapping $M \times M$ windows. Each window’s relative rankings dictate routing: severely degraded regions are processed by the ASP branch, and others by a lightweight convolutional stack. Reconstruction images are reassembled and refined through pointwise convolution within the MFE module, which utilizes stochastic path dropping to maintain coherence and avoid over-smoothing (see Appendix for details). This design enables spatially adaptive restoration, dynamically modulating computational and representational capacity according to local degradation complexity, thereby achieving superior perceptual fidelity.

\begin{algorithm}[H]
\caption{Spatial Structure Adjustment}
\label{al2}
\begin{algorithmic}[1]
\Require Input token features $\{\mathbf{x}_i\}_{i=1}^L$, MSPS graph $\mathcal{G}_H$, transformation matrices $\{\mathbf{U}_i, \mathbf{D}_i\}$
\Ensure Output enhanced features $\{\mathbf{y}_i\}_{i=1}^L$
\State \textbf{Upward pass (Detail Aggregation):}
\For{each node $i$ in bottom-up order}
  \[
    \boldsymbol{\xi}_i = \mathbf{U}_i\Bigl(\mathbf{x}_i + \sum_{j\in\mathcal{N}(i)} \alpha_{ij}\,\boldsymbol{\xi}_j\Bigr),
    \quad \alpha_{ij} = \mathrm{softmax}(w_{ij})
  \]
\EndFor
\State \textbf{Downward pass (Context Refinement):}
\State Initialize $\mathbf{h}_{\mathrm{root}} = \boldsymbol{\xi}_{\mathrm{root}}$
\For{each node $i$ in top-down order}
  \[
    \mathbf{y}_i = \mathrm{Conv}_{1\times1}\bigl[\boldsymbol{\xi}_i \| \mathbf{D}_i\,\mathbf{h}_{\mathrm{par}(i)}\bigr]
  \]
\EndFor
\end{algorithmic}
\end{algorithm}
\subsection{Adaptive Sparse Processing Module}
\label{subsec:dahp}
To address severe and anisotropic degradation, we introduce the ASP module, which constructs an eight-directional, multi-scale propagation hierarchy that selectively routes information along dominant structural trajectories, enabling efficient and structure-aware restoration.

\subsubsection{Structure-Guided Graph Construction}
Each image patch $\mathbf{p}$ is represented by $L$ token nodes, where each node $i$ connects to eight neighbors $j \in \mathcal{N}(i)$ with edge weights jointly encoding spatial and feature similarity:
\begin{equation}
w_{ij} = \exp\bigl(-\beta \|\mathbf{p}_i - \mathbf{p}_j\|_2^2\bigr)
+ \lambda\,\cos(\mathbf{z}_i,\mathbf{z}_j),
\end{equation}
where $\mathbf{p}_i$ and $\mathbf{z}_i$ denote the spatial coordinate and deep feature of node $i$, respectively, and $\beta$, $\lambda$ control the balance between spatial proximity and appearance affinity. To capture both fine details and cross-scale structures with $\mathcal{O}(n)$ complexity, we construct the Minimal Spanning Propagation Structure (MSPS) by pruning with Contractive Borůvka algorithm~\cite{mariano2015generic}, which forms a two-level sparse hierarchy: lower-level edges preserve local detail, while higher-level connections encode global structure.

\subsubsection{Bidirectional Feature Aggregation}
To address the imbalance between fine-detail recovery and global structural consistency in conventional unidirectional propagation, we design a Bidirectional Feature Aggregation (BFA) scheme within MSPS. It integrates bottom-up detail aggregation and top-down contextual refinement into a unified Spatial Structure Adjustment (SSA) mechanism (Algorithm~\ref{al2}), harmonizing local precision with global coherence. Specifically, $\mathbf{U}_i$ gathers direction-aware high-frequency cues, $\mathbf{D}_i$ conveys contextual semantics from parent nodes, and “$|$” denotes feature concatenation for hierarchical interaction between detail restoration and structure preservation.

\subsubsection{Spatial Structure Adjustment Block (SSAB)}
To further modulate the enhanced features, we design a Spatial Structure Adjustment (SSA) Block, defined as:
\begin{equation}
\mathbf{y''} = \mathrm{LN}\bigl(\mathrm{SSA}(\sigma(\mathrm{Conv}(\mathrm{LN}(x'')))) \cdot \sigma(\mathrm{LN}(x''))\bigr),
\end{equation}
where $\mathrm{LN}$ denotes Layer Normalization, $\mathrm{SSA}$ refers to the SSA mechanism, $\sigma$ is the sigmoid activation function, and $x''$ is the input feature map $\mathbf{FM}$. In the ASP module, we stack $n$ SSABs with residual connections:
\begin{equation}
\mathbf{y'} = \mathrm{SSAB}_n\bigl(\mathrm{Norm}(\mathbf{FM})\bigr) + \mathbf{FM},
\end{equation}
where $\mathrm{Norm}$ is a standard normalization operation~\cite{lugmayr2020srflow}. In summary, the ASP module attains substantial complexity reduction by transforming indiscriminate dense computation into structured, content-aware sparsity. Rather than exhaustively modeling all token interactions, ASP constructs a structure-guided sparse graph and derives MSPS, through which information flows only along the most informative trajectories. This design compresses the quadratic interaction burden of global attention into a near-linear propagation process. Working in concert with the PDR module, ASP further activates intensive computation solely within severely degraded regions, while efficiently routing lightly degraded areas through compact convolutional paths—achieving a globally balanced allocation of resources. Within the sparse hierarchy, bidirectional aggregation—bottom-up for fine detail encoding and top-down for contextual reinforcement—preserves rich representational capacity under linear complexity. In essence, ASP redefines efficiency as adaptive selectivity: replacing indiscriminate dense computation with degradation-aware sparse reasoning, thereby realizing a powerful trade-off between computational efficiency and perceptual fidelity.

\begin{table*}[t]
\centering
\caption{We benchmark state-of-the-art super-resolution methods using average PSNR/SSIM, parameter count, and Multiply-Adds (MAdds), computed under a $1280 \times 720$ upscaling setting. Bold and underlined values denote the best and second-best results, respectively. ``\colorbox{cyan!15}{Rel. Imp}" denotes the relative improvements compared with the previous SOTA method.}

\resizebox{\textwidth}{!}{
\renewcommand{\arraystretch}{1.1}
\begin{tabular}{c|c|cc|cc|cc|cc|cc|cc}
\hline
\multirow{2}{*}{Method} & \multirow{2}{*}{Scale} & \multirow{2}{*}{\#Params} & \multirow{2}{*}{\#MAdds} &  \multicolumn{2}{c|}{Set5}   & \multicolumn{2}{c|}{Set14}   & \multicolumn{2}{c|}{BSD100}  &\multicolumn{2}{c|}{Urban100}   & \multicolumn{2}{c}{Manga109}  \\ \cline{5-14}

 &  &  &  & PSNR & SSIM & PSNR & SSIM & PSNR & SSIM & PSNR & SSIM & PSNR & SSIM  \\
\hline
SESR~\cite{bhardwaj2022collapsible} & $ \times 4 $ & 729K & 36.9G& 31.54 & 0.8866& 28.12& 0.7712 & 27.31&0.7277&25.31 &0.7604&29.04&0.8901 \\
IMDN~\cite{hui2019lightweight} & $ \times 4 $ & 715K & 40.9G & 32.21 & 0.8948 & 28.58 & 0.7811 & 27.56 & 0.7353& 26.04 & 0.7838 & 30.45 & 0.9075 \\
LatticeNet~\cite{luo2020latticenet} & $ \times 4 $ & 777K & 43.6G & 32.18 & 0.8943 & 28.61 & 0.7812 & 27.57 & 0.7355 & 26.14 & 0.7844& - & -\\
FDIWN~\cite{gao2022feature} & $ \times 4 $ & 664K & \underline{28.4G} & 32.23 & 0.8955 & 28.66 & 0.7829 & 27.62 & 0.7380 & 26.28 & 0.7919 & - & -\\
SwinIR-light~\cite{liang2021swinir} & $ \times 4 $ & 930K & 61.7G & 32.44 & 0.8976 & 28.47 & 0.7823 & 27.57 & 0.7402 & 26.47 & 0.7980 & 30.92 & 0.9051\\
SwinFIR~\cite{zhang2022swinfir} & $ \times 4 $ & 901K & 43.2G & 32.43 & 0.8975 & 28.48 & 0.7849 & 27.59 & 0.7406 & 26.54 & 0.7982 & 30.92 & 0.9050\\
NGSwin~\cite{choi2023n} & $ \times 4 $ & 719K & 36.4G & 32.33 & 0.8963 & 28.74 & 0.7853 & 27.66 & 0.7396  & 26.45 & 0.7963 & 30.80 & 0.9128 \\
SwinIR-NG~\cite{choi2023n} & $ \times 4 $ & 1010K & 63.0G & 32.44 & 0.8980 & 28.63 & 0.7870 & 27.71 & 0.7411 & 26.54 & 0.7998 & 31.09 & 0.9061\\
DiVANet~\cite{behjati2023single} & $ \times 4 $ & 839K & 57.0G & 32.41 & 0.8973 & 28.70 & 0.7844 & 27.65 & 0.7391 & 26.42 &  0.7958 & 30.73 & 0.9119\\
PACN~\cite{wang2023pixel} & $ \times 4 $&15M &82.9G& 32.42 &0.8975 &28.56 &0.7856 &27.66&0.7388 & 26.49 &0.8007 & 31.02 &0.9156 \\
CVANet~\cite{zhang2024cvanet}  & $ \times 4 $& 1232K & 52.8G & 32.15 &0.8965 &28.69 &0.7864 & 27.59 & 0.7408 & 26.57 & 0.8008 & 31.13 &0.9152\\
SSIR~\cite{zhao2024ssir}& $ \times 4 $&7368K & 79.5G&32.48 &0.8925 & 28.80 &0.7864 &27.68&0.7371 & 26.39 &0.8001 &31.08 &0.9146\\
MAN~\cite{wang2024multi}& $ \times 4 $& 8.5M & 420G & 32.49 &0.8983 &28.77 &0.7867 & 27.71&0.7404 &26.58 &0.8009 &31.14 &0.9157\\
DVMSR~\cite{lei2024dvmsr} & $ \times 4 $& \underline{544K} &42.8G &32.17&0.8951&28.58&0.7823&27.57&0.7378& 26.01&0.7836&30.51&0.9083\\
CAMixerSR ~\cite{wang2024camixersr} & $ \times 4 $ & 765K & 53.8G & 32.51 & 0.8975 & \underline{28.82} & \underline{0.7873} & 27.72 & 0.7416 & 26.54 & 0.7982  & 30.92 &0.9150 \\
HASN~\cite{cao2025hasn} & $\times 4 $ & 932K & 60.2G & 32.44 & 0.8947 &28.76 &0.7844 &27.47 &0.7413 &26.44 & 0.7988&30.69&0.9123\\
SRMamba-T~\cite{liu2025srmamba}& $ \times 4 $& 681K &29.5G &32.33 &0.8968&28.76&0.7859&27.52&0.7397&26.53&0.8004&30.99&0.9138\\
ASID~\cite{park2025efficient} & $ \times 4 $& 920K &48.5G &\underline{32.54}&0.8945&28.52&0.7833 &\underline{27.77} &\underline{0.7436} &26.32&0.8003&31.16 &0.9123\\
SpikeSR~\cite{xiao2025spiking} & $\times 4 $ & 789K &46.8G & 32.52 & \underline{0.8988}&28.78 &0.7858 & 27.69 & 0.7406 &\underline{26.63} & \underline{0.8012}&\underline{31.18} & \underline{0.9166}\\ \hline
 \rowcolor{red!15} DPAMixerSR  & $ \times 4 $ &  \textbf{383K} &  \textbf{24.7G} & \textbf{33.87} & \textbf{0.9148} & \textbf{30.09} & \textbf{0.8102} & \textbf{28.95} & \textbf{0.7663} & \textbf{27.97} & \textbf{0.8125} & \textbf{32.86} & \textbf{0.9298}\\

\rowcolor{cyan!15} Rel. Imp & - & -161K &-3.7G & 1.33 & 0.0160 & 1.27 &  0.0229 & 1.18 & 0.0227 & 1.34 &0.0113 &1.68 &0.0132\\

\hline
\end{tabular}
}
\label{sota}
\end{table*}

\section{Experiments}
\begin{table*}[ht]
\centering
\caption{Quantitative comparison with state-of-the-art Real-ISR methods on four real-world benchmarks. Best and second best performance are highlighted in  \colorbox{red!15}{red} and  \colorbox{cyan!15}{blue}, respectively.}
\resizebox{\textwidth}{!}{%
\renewcommand{\arraystretch}{1.1}
\begin{tabular}{l|c|c|c|c|c|c|c|c|c|c|c|c}
\hline
\multirow{2}{*}{Datasets} & \multirow{2}{*}{Metrics} & \multicolumn{11}{c}{Methods} \\ \cline{3-13}  
 &  & MAT~\cite{xie2025mat} & FoundIR~\cite{li2025foundir} & ResShift~\cite{yue2023resshift} & StableSR~\cite{wang2024exploiting} & SeeSR~\cite{wu2024seesr} & DiffBIR~\cite{lin2024diffbir} & OSEDiff~\cite{wu2024one} & SUPIR~\cite{yu2024scaling} & TVT~\cite{yi2025fine}  & DiT4SR~\cite{duan2025dit4sr} & \textbf{DPAMixerSR} \\ \hline
\multirow{5}{*}{DrealSR} & LPIPS $\downarrow$ & 0.282 & 0.274 & 0.353 & \colorbox{cyan!15}{0.273} & 0.317 & 0.452 & 0.297 & 0.419 & 0.354  & 0.365 &\colorbox{red!15}{\textbf{0.259}}\\ 
 & MUSIQ $\uparrow$ &\colorbox{cyan!15}{65.665}  & 52.737 & 52.392 & 58.512 & 65.077 & 57.341 & 64.692 & 59.744 & 44.047 &  64.950 &\colorbox{red!15}{\textbf{66.841}}\\ 
 & MANIQA $\uparrow$ & 0.490 & 0.475 & 0.476 & 0.559 & 0.605 & 0.455  & 0.590 & 0.552 & \colorbox{cyan!15}{0.631}&  0.627 & \colorbox{red!15}{\textbf{0.645}}\\ 
 & ClipIQA $\uparrow$ & 0.401 & 0.396 & 0.379 & 0.438 & 0.543 & 0.479 & 0.519 & 0.512 & \colorbox{cyan!15}{0.576} &  0.548 & \colorbox{red!15}{\textbf{0.587}} \\ 
 & LIQE $\uparrow$ & 2.927 & 2.745 & 2.798 & 3.243 & 3.401 & 3.894 & 3.942 & 3.728 &  \colorbox{cyan!15}{4.129} & 3.964 & \colorbox{red!15}{\textbf{4.254}}\\ \hline
\multirow{5}{*}{RealSR} & LPIPS $\downarrow$ & 0.271 & \colorbox{cyan!15}{0.254} & 0.316 & 0.306 & 0.299 & 0.347 & 0.292 & 0.357 & 0.325 &  0.319 &\colorbox{red!15}{\textbf{0.241}}\\ 
 & MUSIQ $\uparrow$ & 60.370 & 58.694 & 56.892 & 65.653 & \colorbox{cyan!15}{69.675} & 68.340 & 69.087 & 61.929 & 59.396 &  68.073 &\colorbox{red!15}{\textbf{70.148}}\\ 
 & MANIQA $\uparrow$ & 0.551 & 0.524 & 0.511 & 0.622 & 0.643 & 0.653 & 0.634 & 0.574 & 0.546 &  \colorbox{cyan!15}{0.661} & \colorbox{red!15}{\textbf{0.678}} \\ 
 & ClipIQA $\uparrow$ & 0.432 & 0.422 & 0.407 & 0.472 & 0.577 & \colorbox{cyan!15}{0.586} & 0.552 & 0.543 & 0.474 &  0.550 & \colorbox{red!15}{\textbf{0.603}}\\ 
 & LIQE $\uparrow$ & 3.358 & 2.956 & 2.853 & 3.750 &  \colorbox{cyan!15}{4.123} & 4.026 & 4.065 & 3.780 & 3.221 &  3.977 & \colorbox{red!15}{\textbf{4.317}}\\ \hline
\multirow{4}{*}{RealLR200} & MUSIQ $\uparrow$ & 62.961 & 63.548 & 59.695 & 63.433 & 69.428 & 68.027 & 69.547 & 64.837 & 65.926 &   \colorbox{cyan!15}{70.469} & \colorbox{red!15}{\textbf{72.461}} \\ 
 & MANIQA $\uparrow$ & 0.553 & 0.560 & 0.525 & 0.579 & 0.612 & 0.629 & 0.606 & 0.600 & 0.597 &   \colorbox{cyan!15}{0.645} & \colorbox{red!15}{\textbf{0.669}} \\ 
 & ClipIQA $\uparrow$ & 0.451 & 0.463 & 0.452 & 0.458 & 0.566 & 0.582 & 0.551 & 0.524 & 0.546 &   \colorbox{cyan!15}{0.588} & \colorbox{red!15}{\textbf{0.609}} \\ 
 & LIQE $\uparrow$ & 3.484 & 3.465 & 3.054 & 3.379 & 4.006 & 4.003 & 4.069 & 3.626 & 3.775 &   \colorbox{cyan!15}{4.331} & \colorbox{red!15}{\textbf{4.572}} \\ \hline
\multirow{4}{*}{RealLQ250} & MUSIQ $\uparrow$ & 62.514 & 63.371 & 59.337 & 56.858 & 70.556 & 69.876 & 69.580 & 66.016 & 66.693 &  \colorbox{cyan!15}{71.832} & \colorbox{red!15}{\textbf{73.471}} \\ 
 & MANIQA $\uparrow$ & 0.524 & 0.534 & 0.500 & 0.504 & 0.594 & 0.624 & 0.578 & 0.584 & 0.585 &  \colorbox{cyan!15}{0.632} & \colorbox{red!15}{\textbf{0.647}} \\ 
 & ClipIQA $\uparrow$ & 0.435 & 0.440 & 0.417 & 0.382 & 0.562 & \colorbox{cyan!15}{0.578} & 0.528 & 0.483 & 0.502 &  \colorbox{cyan!15}{0.578} & \colorbox{red!15}{\textbf{0.598}}\\ 
 & LIQE $\uparrow$ & 3.341 & 3.280 & 2.753 & 2.719 & 4.005 & 4.003 & 3.904 & 3.605 & 3.688 &  \colorbox{cyan!15}{4.356} & \colorbox{red!15}{\textbf{4.562}} \\ \hline
\end{tabular}%
}
\label{realworld}
\end{table*}

\begin{figure}[htbp]
  \centering
  \includegraphics[width=0.8\linewidth]{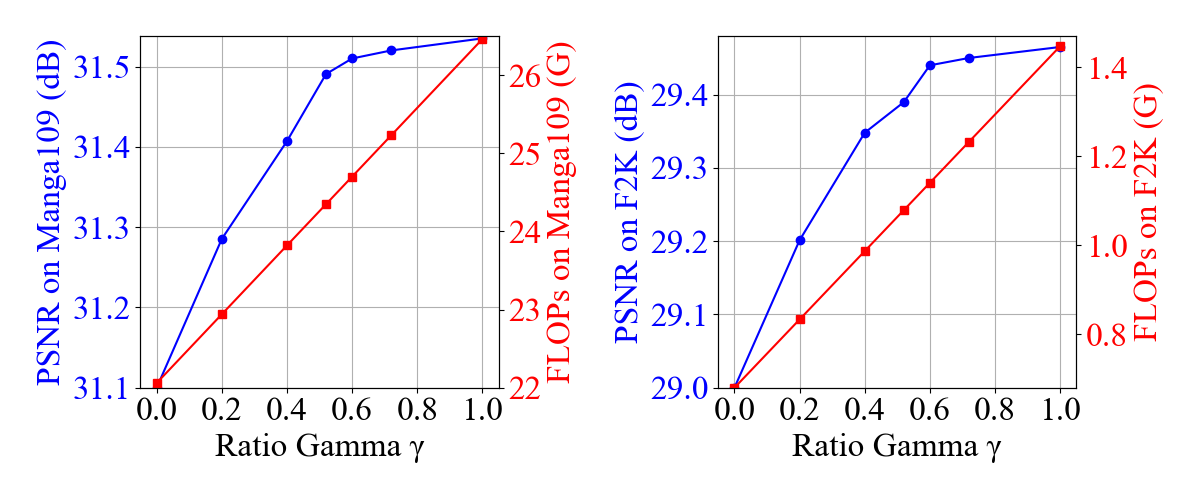}
  \caption{Ablation study on ASP ratio $\gamma$.} 
  \setlength{\belowcaptionskip}{-3mm}
  \label{visual2}
\end{figure}

\subsection{Implementation Details}
\textbf{Experimental Setup.} Following ~\cite{zhang2022efficient,wang2024camixersr} , our backbone consists of 20 DPAMixer and FFN blocks, with 60 channels. The PDR mechanism operates within a window size of 16, while the convolutional branch incorporates two 3×3 depth-wise convolutions. We tuned $\epsilon$ over $\{0.05, 0.1, 0.15\}$ and found that $\epsilon = 0.1$ yields the best performance.

\noindent\textbf{Datasets and Evaluation Metrics.}
For PDR, we use 140K pristine images from dataset KADIS~\cite{lin2019kadid} for unsupervised training, and  CSIQ~\cite{larson2010most}, and TID2013~\cite{ponomarenko2013color} for supervised learning. We train on DIV2K~\cite{agustsson2017ntire}, evaluating methods on Set5~\cite{wang2024camixersr}, Set14~\cite{zeyde2012single}, BSD100~\cite{wang2024multi}, Urban100~\cite{lei2024dvmsr}, and Manga109~\cite{matsui2017sketch}.

\noindent\textbf{Baselines.} We compare DPAMixerSR with various SOTA SR models, including some Transformer-based models (SSIR~\cite{zhao2024ssir}, SwinIR-light~\cite{liang2021swinir}, SwinFIR~\cite{zhang2022swinfir}, NGSwin~\cite{choi2023n}, SwinIR-NG~\cite{choi2023n}, PACN~\cite{wang2023pixel}, CVANet~\cite{zhang2024cvanet}, CAMixerSR~\cite{wang2024camixersr}, HASN~\cite{cao2025hasn}, ASID~\cite{park2025efficient} and SpikeSR~\cite{xiao2025spiking}), CNN-based models ( IMDN~\cite{hui2019lightweight}, LatticeNet~\cite{luo2020latticenet}, FDIWN~\cite{gao2022feature}, SESR~\cite{bhardwaj2022collapsible}, and DiVANet~\cite{behjati2023single}), and 
Mamba-based models(MAN~\cite{wang2024multi}, DVMSR~\cite{lei2024dvmsr} and SRMamba-T~\cite{liu2025srmamba}).

\subsection{Lightweight SR}
Table~\ref{sota} demonstrates DPAMixerSR's superiority over state-of-the-art lightweight SR methods. Our approach achieves the highest PSNR/SSIM across all datasets while maintaining significantly lower computational complexity. Its degradation-aware design enables superior reconstruction of complex textures in Urban100 (dense patterns with compound degradations) and Manga109 (topology-sensitive line art). Through structure-guided graph construction and bidirectional feature aggregation, we accurately restore Urban100's fine textures while preserving Manga109's long-range structural coherence—particularly across challenging double-page layouts where conventional convolutions fail.

\subsection{Real-World Super-Resolution}
To further validate the scalability and generalization of DPAMixerSR, we evaluate it on real-world SR tasks characterized by unknown and spatially variant degradations that pose challenges beyond synthetic settings. Experiments are conducted on four widely used datasets: DRealSR~\cite{wei2020component}, RealSR~\cite{cai2019toward}, RealLR200~\cite{wu2024seesr}, and RealLQ250~\cite{ai2024dreamclear}. Given full-reference metrics like PSNR and SSIM often misalign with human perception~\cite{blau2018perception,jinjin2020pipal}, we primarily report LPIPS~\cite{zhang2018unreasonable}, complemented by user studies and no-reference metrics such as MUSIQ~\cite{ke2021musiq}, MANIQA~\cite{yang2022maniqa}, ClipIQA~\cite{wang2023exploring}, and LIQE~\cite{zhang2023blind}.
We compare against representative real-world SR baselines~\cite{xie2025mat,li2025foundir,yue2023resshift,wang2024exploiting,wu2024seesr,lin2024diffbir,wu2024one,yu2024scaling,yi2025fine,duan2025dit4sr}. 
In Table~\ref{realworld}, DPAMixerSR achieves the best perceptual performance across all benchmarks. These results demonstrate DPAMixerSR's ability to produce high-quality restorations.

\subsection{Ablation Study}

\begin{figure}[htbp]
  \centering
  \includegraphics[width=0.8\linewidth]{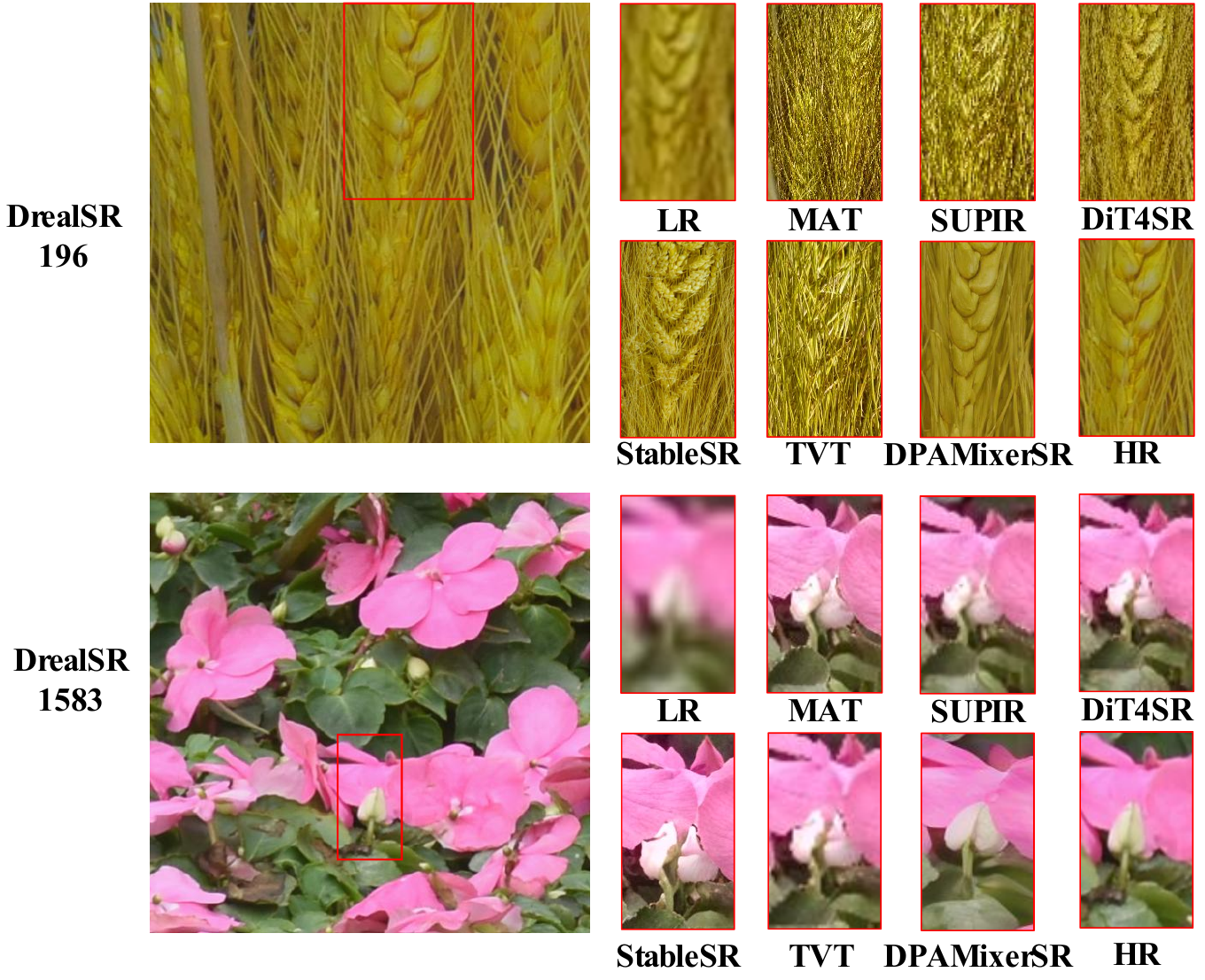}
  \caption{Qualitative comparison on the real-world dataset.} 
  \label{visual_realworld}
\end{figure}

\noindent\textbf{Visual Comparison and Analysis.}
Figure~\ref{visual_realworld} demonstrates consistent gains on real-world scenes with complex degradations, such as motion-blurred wheat fields and densely textured floral images. DPAMixerSR reconstructs sharper details and more coherent structures—e.g., distinct awns, petal veins, and bud contours—benefiting from precise patch-level diagnosis via the Perceptual Degradation Assessment module and efficient sparse propagation within ASP.

\begin{table}[h]
\centering
\caption{Ablation studies on our method on RealSR dataset.}
\label{tab:ablation_tvt_realisr}
\resizebox{0.8\textwidth}{!}{
\begin{tabular}{l|rrrrrr}
\hline
Methods & FLOPs & LPIPS$\downarrow$ & MUSIQ$\uparrow$ & MANIQA$\uparrow$ & ClipIQA$\uparrow$ & LIQE$\uparrow$\\
\hline
TVT~\cite{yi2025fine} &1.97 T & 0.325 & 59.396  & 0.546 & 0.474 & 3.221  \\ \hline
DPAMixerSR (w/o PDR)& 101 G& 0.276 &65.573 & 0.624 & 0.569 & 4.047 \\
DPAMixerSR (w/o PDR, w/ PDS)& 285 G& 0.234 &71.462 & 0.683 & 0.618 & 4.330 \\
DPAMixerSR (w/o SSA)& 68.9 G& 0.295 &67.463 & 0.653 & 0.586 & 4.171 \\
DPAMixerSR (w/o SSA, w/ TSP)& 214 G& 0.283 &67.821 & 0.659 & 0.590 & 4.178 \\ 
DPAMixerSR (w/o SSA, w/ EIT-GNN)& 259 G& 0.287 &67.514 & 0.643 & 0.573 & 4.153 \\ 
 DPAMixerSR& 128 G& 0.241 &70.148 & 0.678 & 0.603 & 4.317 \\
\hline
\end{tabular}
}
\end{table}

\noindent\textbf{Effectiveness of Each Component.}
To systematically dissect the individual functional modules embedded in our proposed DPAMixerSR framework and quantify their respective performance contributions, we perform extensive ablation experiments whose quantitative results are summarized in Table~\ref{tab:ablation_tvt_realisr}. In this ablation analysis, we separately remove or substitute each core building block, including the PDR module, PDS module, and SSA module, and further incorporate state-of-the-art sparse graph modeling paradigms TSP~\cite{zhou2025tsp} and EIT-GNN~\cite{park2024graph} as competitive alternative baselines for cross-comparison. Experimental observations reveal that every proposed component brings steady and measurable improvements across all evaluation metrics. Such consistent performance increments collectively validate that PDR, PDS and SSA serve mutually complementary roles in capturing spatial dependencies and refining high-resolution feature representations, and the integration of all modules jointly delivers the optimal reconstruction performance of DPAMixerSR.

\section{Conclusion}
In this work, we present DPAMixerSR, a unified framework for efficient image super-resolution based on degradation-adaptive computation. To this end, we introduce a Perceptual Degradation Assessment (PDA) module that quantifies restoration difficulty and enables dynamic routing guided by degradation semantics. We further design an Adaptive Sparse Processing (ASP) module that performs sparse hierarchical propagation and bidirectional refinement along dominant structures, achieving near-linear complexity under anisotropic degradations.
Extensive experiments demonstrate that DPAMixerSR attains state-of-the-art perceptual quality with substantially lower computational cost, exhibiting strong generalization to large-scale and real-world SR tasks. Beyond image SR, the framework provides a natural foundation for extending degradation-adaptive computation to video restoration and other resource-constrained vision tasks.

\section*{Acknowledgements}
This work was supported by National Natural Science Foundation of China (Grant No. U25A20537).

%
%
%
\bibliographystyle{splncs04}
\bibliography{mybibliography}
%




\end{document}